\documentclass[sigconf]{acmart}

\copyrightyear{2026}
\acmYear{2026}
\setcopyright{cc}
\setcctype{by}
\acmConference[WWW Companion '26] {Companion Proceedings of the ACM Web Conference 2026}{April 13--17, 2026}{Dubai, United Arab Emirates.}
\acmBooktitle{Companion Proceedings of the ACM Web Conference 2026 (WWW Companion '26), April 13--17, 2026, Dubai, United Arab Emirates}
\acmISBN{979-8-4007-2308-7/2026/04}
\acmDOI{10.1145/3774905.3793119}

\usepackage{multirow}

\begin{document}

\title{CiteLLM: An Agentic Platform for Trustworthy Scientific Reference Discovery}

\author{Mengze Hong}
\author{Di Jiang}
\authornote{Corresponding Author}
\affiliation{%
  \institution{Hong Kong Polytechnic University}
  \city{Hong Kong}
  \country{China}
}

\author{Chen Jason Zhang}
\author{Zichang Guo}
\affiliation{%
  \institution{Hong Kong Polytechnic University}
  \city{Hong Kong}
  \country{China}
}

\author{Yawen Li}
\affiliation{%
  \institution{Beijing University of Posts and Telecommunications}
  \city{Beijing}
  \country{China}
}

\author{Jun Chen}
\affiliation{%
  \institution{Independent Researcher}
  \city{Hong Kong}
  \country{China}
}

\author{Shaobo Cui}
\affiliation{%
  \institution{Swiss Federal Technology Institute of Lausanne (EPFL)}
  \city{Lausanne}
  \country{Switzerland}
}

\author{Zhiyang Su}
\affiliation{%
  \institution{Hong Kong University of Science and Technology (HKUST)}
  \city{Hong Kong}
  \country{China}
}

\renewcommand{\shortauthors}{Mengze Hong et al.}

\begin{abstract}
Large language models (LLMs) have created new opportunities to enhance the efficiency of scholarly activities; however, challenges persist in the ethical deployment of AI assistance, including (1) the trustworthiness of AI-generated content, (2) preservation of academic integrity and intellectual property, and (3) protection of information privacy. In this work, we present CiteLLM, a specialized agentic platform designed to enable trustworthy reference discovery for grounding author-drafted claims and statements. The system introduces a novel interaction paradigm by embedding LLM utilities directly within the LaTeX editor environment, ensuring a seamless user experience and no data transmission outside the local system. To guarantee hallucination-free references, we employ dynamic discipline-aware routing to retrieve candidates exclusively from trusted web-based academic repositories, while leveraging LLMs solely for generating context-aware search queries, ranking candidates by relevance, and validating and explaining support through paragraph-level semantic matching and an integrated chatbot. Evaluation results demonstrate the superior performance of the proposed system in returning valid and highly usable references.
\end{abstract}



\maketitle

\section{Introduction}

In academic writing, grounding claims and arguments in reliable scientific sources is essential for maintaining scholarly rigor \cite{national2011reference}. Unlike literature surveys or related work sections, which broadly summarize prior research on a topic, reference discovery focuses on identifying precise materials that support or refute a specific statement or finding. This task is more challenging, as it requires careful consideration of the target statement, its surrounding context, and the overall goals of the manuscript. Traditionally, this process involves repeated searches on academic repositories such as Google Scholar using diverse keywords, followed by reviewing numerous papers to identify the most relevant ones, making it highly time-consuming. This has motivated AI-driven automation and attracted significant interest from the NLP and web communities \cite{10.1145/3701716.3735083, 10.1145/3487553.3524657}.

\begin{figure}[!t]
    \centering
    \includegraphics[width=0.85\linewidth]{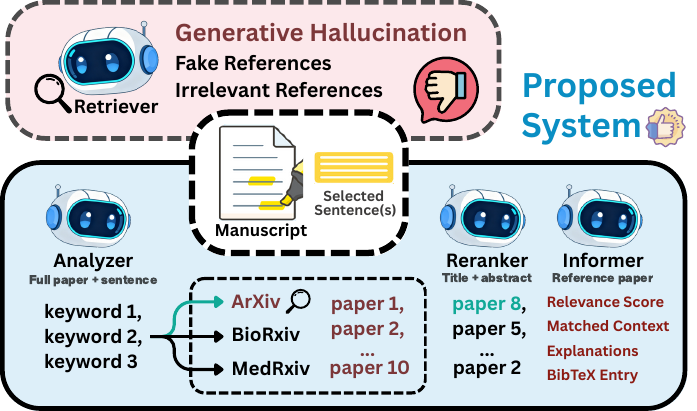}
    \vspace{-0.5em}
    \caption{Overview of CiteLLM: trustworthy reference discovery powered by three LLM-based utility agents.}
    \label{fig:overview}
    \vspace{-1em}
\end{figure}

Large language models (LLMs) are increasingly employed to draft, revise, and refine research articles~\cite{chen2025ai4researchsurveyartificialintelligence}. However, a major risk of AI-assisted writing lies in trustworthiness: incorrect, fabricated, or poorly supported content can undermine scholarly integrity and violate emerging ethical guidelines and venue policies that warn against unverified generative outputs and non-privacy-preserving workflows~\cite{naddaf2025ai,elali2023ai,checco2021ai}. This issue is particularly pronounced in reference discovery, where fake or irrelevant literature introduced by LLM hallucinations can severely weaken arguments. At the same time, serious data-privacy concerns arise, as uploading unpublished manuscripts to third-party LLMs or citation services risks leaking sensitive pre-publication research, limiting the practical use of existing tools such as Google Scholar Labs.

\begin{figure*}[!t]
  \centering
  \includegraphics[width=0.82\linewidth]{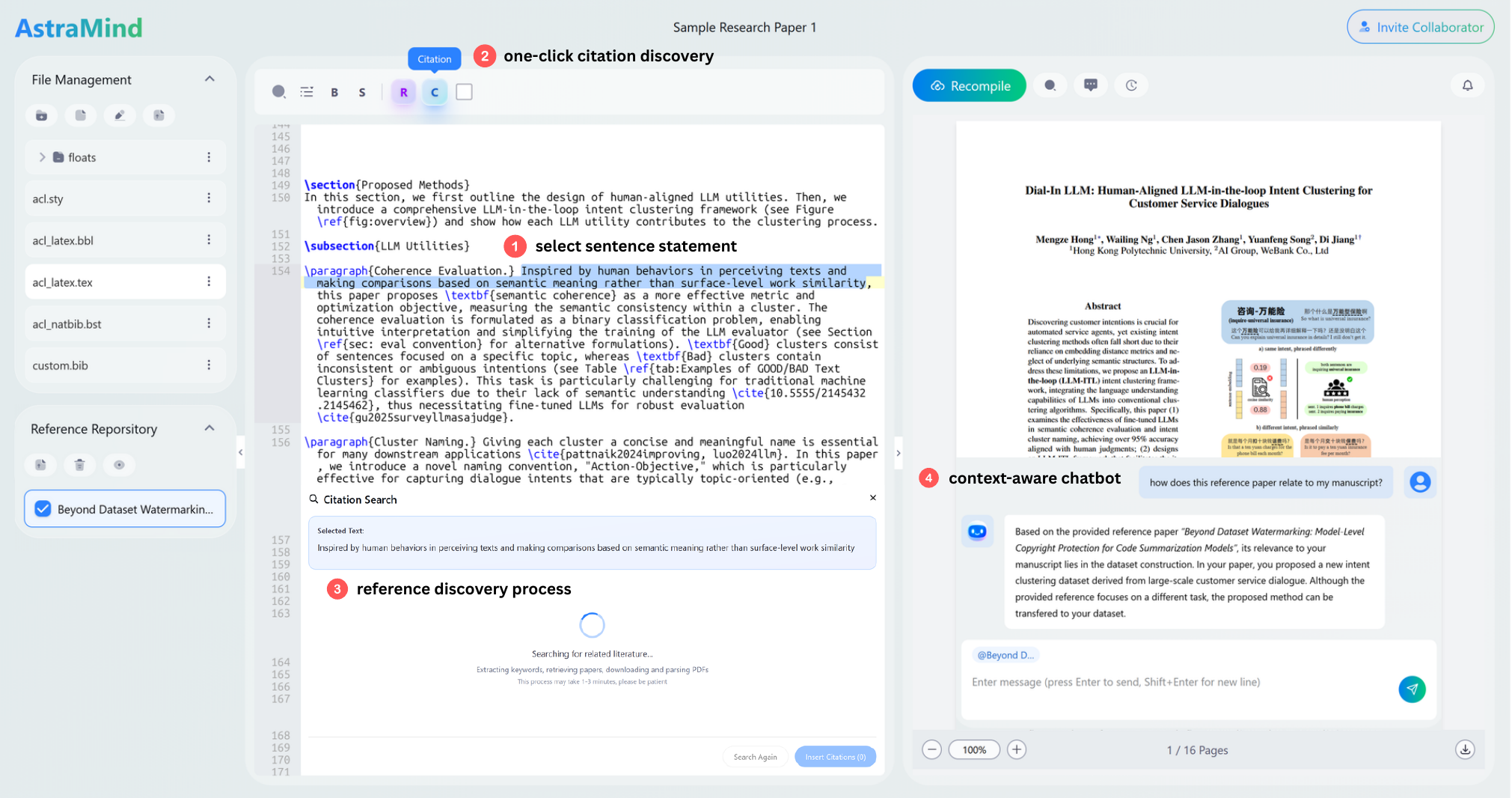}
  \caption{Demonstration of user interface: a full LaTeX editor with embedded reference discovery in four steps: (1) select sentence, (2) one-click citation discovery, (3) reference search process, (4) context-aware chatbot discussion}
  \vspace{-1em}
  \label{fig:citation-ui}
\end{figure*}

In this paper, we introduce \emph{\textbf{CiteLLM}}, a specialized agentic platform designed to support trustworthy reference discovery with verifiable prior literature (see Figure~\ref{fig:overview}). By focusing on strengthening the provenance of author-written statements within an integrated system environment, rather than generating new scientific content, the system upholds ethical research practices throughout the academic writing process. Our core contributions are threefold:

\begin{enumerate}
    \item A novel interactive interface that integrates LLM assistance into the LaTeX editing environment, allowing authors to select a sentence or short text span directly within their manuscript to retrieve relevant references, while automatically incorporating surrounding and document-level context;
    \item A fully automated reference discovery pipeline that generates precise, context-aware queries, retrieves candidate papers, verifies relevance through document parsing, and delivers a ranked list of ready-to-use citations with human-verifiable evidence and a one-click citation insertion utility;
    \item A dynamic routing mechanism that selectively queries multiple trusted web-based academic repositories, fundamentally eliminating the risk of LLM-generated references while ensuring comprehensive coverage of relevant literature.
\end{enumerate}

\section{Proposed System}

\subsection{User Interaction Design}

To streamline academic writing workflow while enforcing strict data sovereignty and ethical LLM deployment, the proposed system is natively embedded within the LaTeX editor environment, ensuring that all manuscript content and intermediate representations remain local and are never transmitted to external servers. Authors compose and compile in real-time within a single, self-contained workspace. We assert that \textbf{optimal AI integration occurs precisely at the point of authorship}, formalized as co-localizing the assistance utility $f_{\text{AI}}(\cdot)$ with the editing context $C_t$ at time $t$, thereby eliminating off-site transfer of unpublished material and eliminating external privacy leakage during routine operation.

\begin{figure}
    \centering
    \includegraphics[width=1\linewidth]{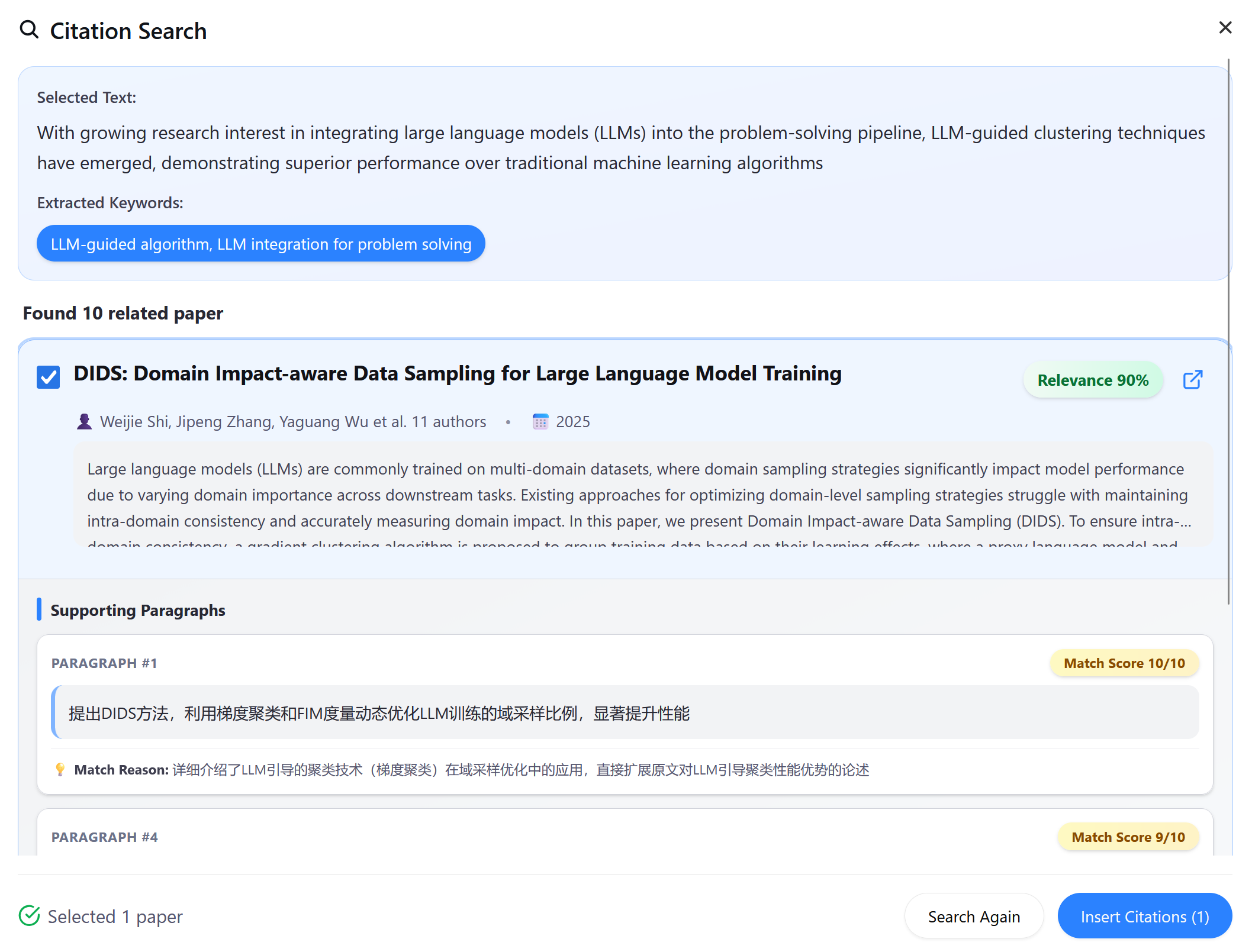}
    \vspace{-2em}
    \caption{Example of reference discovery result.}
    \label{fig:citation-result}
    \vspace{-1.5em}
\end{figure}

Interaction follows a minimal, context-aware paradigm (see Figure \ref{fig:citation-ui}): the author highlights a target claim $S$ from either a sentence or a paragraph span, then invokes the citation discovery module discussed in the following sections. As shown in Figure~\ref{fig:citation-result}, the pipeline returns an ordered set of candidate references $\{(p_1, r_1), (p_2, r_2), \dots \}$, where $p_i$ is a paper record and $r_i \in [0,1]$ its relevance score computed against the selected sentence claim. The top-ranked result $p^* = \arg\max_i r_i$ is accompanied by a natural-language explanation of evidential alignment. With a single click, the system inserts the in-text marker \texttt{\textbackslash cite\{$p^*_{key}$\}} at the end of selected text and appends the corresponding BibTeX entry to ``\texttt{references.bib}'', preserving bibliographic consistency.

To gain further insights about a selected reference after insertion, the system provides an LLM-driven chatbot for natural language interaction. Let \(\mathcal{C}\) denote the full paper context summarized from the LaTeX document, and \(P_j\) the selected reference, with discovery metadata \(\mathcal{M}_j\) containing the target claim \(S_j\) for which \(P_j\) was retrieved. The chatbot is pre-filled with \((\mathcal{C}, P_j, \mathcal{M}_j)\), enabling it to understand both the complete document context and the specific claim linked to the reference. This design provides more accurate, context-aware assistance than external LLM services that lack awareness of the author’s paper, supporting precise expression and iterative manuscript refinement.

\subsection{Context-Augmented Reference Retrieval}

The core utility of the system lies in the reference retrieval module, which employs an intelligent routing strategy with multi-source aggregation to provide precise and reliable literature support for academic writing. Unlike traditional keyword-based matching, this module leverages the advanced semantic understanding capabilities of LLMs for information retrieval \cite{hong2026ral2mretrievalaugmentedlearningtomatch}, complemented by multiple large-scale academic databases, enabling end-to-end automation from user intent recognition to precise literature retrieval.

\textbf{Intelligent Discipline-Based Routing.} Effective scholarly writing requires grounding claims in literature from the target research community (e.g., NLP sources are strongly preferred over computer vision papers for ACL submissions). The system first infers the disciplinary profile of the selected text using a lightweight LLM classifier. To minimize deployment costs and ensure reproducibility, the current implementation relies on public preprint repositories: arXiv (computer science, physics, and mathematics), bioRxiv (biology-related claims), and medRxiv (clinical and medical research). The architecture naturally extends to restricted conference or journal corpora, provided appropriate database or API access is available. When cross-disciplinary characteristics are detected, the system adopts a primary-secondary repository strategy to ensure complete retrieval coverage. The routing decision process employs structured JSON output, containing primary repository (\texttt{primary\_repo}), backup repository (\texttt{secondary\_repo}), confidence level, and reasoning, which provide an explainable basis for subsequent retrieval.

\textbf{Semantically Driven Query Construction:} For text selected by the user, the system applies sentence-level segmentation to identify key claims. The resulting set of sentences is processed in a single batched LLM call to generate relevant keywords, using a prompt that integrates a summarized schema of the full manuscript with the surrounding context. This batch-prompting design reduces latency compared to issuing separate, per-sentence queries, while preserving intersentence context to produce more coherent and accurate results. Query construction follows a priority scheme that first identifies domain-specific technical terms and then integrates core academic concepts grounded in the manuscript context, a process that is well-suited to efficient execution by a locally deployed small LLM, thereby enabling privacy-preserving operation with all research content kept on the local device.

\textbf{Fault Tolerance and BibTeX Acquisition.} Leveraging the routing decision, the system concurrently queries the designated repositories using the constructed queries. If the primary repository returns no results, an automatic fallback to the secondary repository is triggered, substantially increasing retrieval success rates. Results are deduplicated by title, and BibTeX entries are obtained automatically: for arXiv papers via paper ID, and for bioRxiv/medRxiv papers via DOI content negotiation at doi.org. The module further incorporates a retry mechanism with exponential backoff to improve robustness and ensure the delivery of ready-to-use citations.

\subsection{Reference Verification and Usability Support}

Reference verification is essential for citation credibility. Our system employs a two-layer mechanism that utilizes full-text semantic analysis: it first confirms the existence and accessibility, and then rigorously evaluates the semantic alignment between the candidate reference paper and the selected claim. Validated references are further supported by LLM-generated insights that help users understand and effectively integrate the material into the manuscript.

\textbf{Full-Text Acquisition and Parsing.} A critical verification step confirms that each retrieved reference actually exists and is accessible. The system extracts PDF links from retrieval results, prioritizing open-access sources. Full-text processing employs the GROBID tool\footnote{https://github.com/kermitt2/grobid}, specifically designed for scholarly PDFs, which outputs structured TEI XML and accurately preserves section titles, paragraph boundaries, and references. Compared to generic OCR-based tools, GROBID offers significantly higher structural fidelity, ensuring reliable and high-quality input for measuring relevance.

\textbf{Semantic Matching and Context Informer:}
Once the full text is parsed, the system performs fine-grained semantic matching to rank candidate papers in descending order of an overall relevance score computed using a few-shot LLM. As shown in Figure~\ref{fig:citation-result}, each candidate is presented as a structured record containing the title, PDF link, abstract, and the top three matched paragraphs, each with its index, relevance score, and matching rationale. This concise yet comprehensive format allows users to quickly assess relevance, access the full text, or directly insert ready-to-use BibTeX entries and generate in-text citations from the search results, thereby minimizing citation management overhead.

\begin{table*}[!t]
  \centering
  \caption{Performance comparison of citation discovery methods ($N=40$ sentences, top-5 references per sentence).}
  \vspace{-0.5em}
  \label{tab:evaluation}
  \begin{tabular}{lccccc}
    \toprule
    \multirow{2}{*}{\textbf{Method}} & \multirow{2}{*}{Validity (\%)} & \multicolumn{2}{c}{Precision (\%)} & \multicolumn{2}{c}{Usability (\%)} \\
    \cmidrule(lr){3-4} \cmidrule(lr){5-6}
    & & Human & GPT-5 & Human & GPT-5 \\
    \midrule
    \textbf{CiteLLM (ours)} & \textbf{100} & \textbf{84.0} & \textbf{87.5} & \textbf{87.5} & \textbf{92.5} \\
    Google Scholar & 100 & 64.0 & 73.5 & 80.0 & 85.0 \\
    *ChatGPT & 74 & 84.5 & 91.9 & 30.0 & 37.5 \\
    \bottomrule
  \multicolumn{6}{l}{\footnotesize{*Note: Precision for ChatGPT is evaluated among the valid references.}}
  \vspace{-0.5em}
  \end{tabular}
\end{table*}

\section{Experiments}
We evaluate the proposed system using 40 sentence statements sampled from publicly available research papers across multiple disciplines, each accompanied by human-annotated search queries and at least one ground-truth reference from its original paper.

\subsection{Query Construction}

\begin{figure}
    \centering
    \includegraphics[width=1\linewidth]{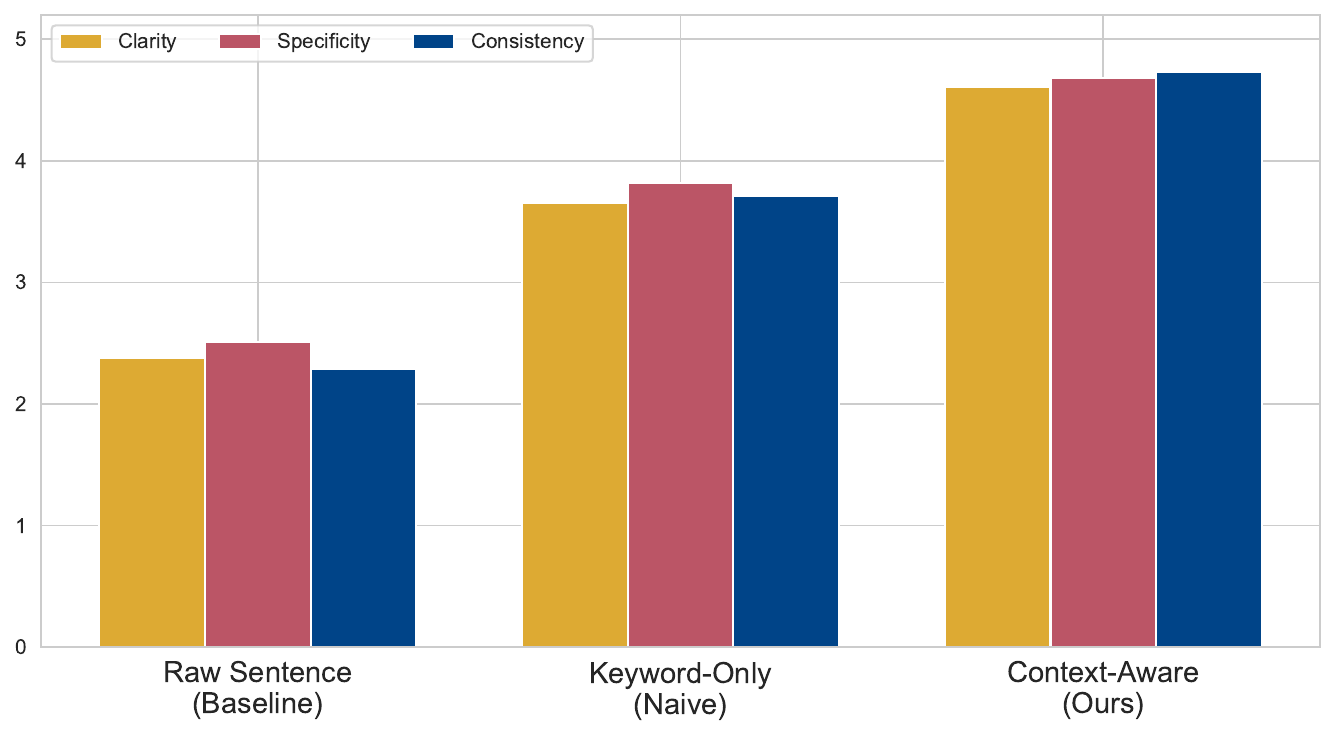}
    \vspace{-2.5em}
    \caption{Comparison of query construction approaches.}
    \label{fig:query-quality}
    \vspace{-1.5em}
\end{figure}

We first examine how contextual information improves the quality of reference search queries. Three query variants are compared: (1) a baseline using the raw segmented sentence alone, (2) a naive keyword-only query extracted solely from the sentences via a standard LLM prompt, and (3) the proposed context-aware query that combines extracted keywords with manuscript context. Three experienced researchers independently score each query on a 5-point Likert scale for clarity, specificity, and consistency with human-annotated search queries. As shown in Figure~\ref{fig:query-quality}, the proposed context-aware approach consistently outperforms both baselines across all three dimensions, demonstrating its ability to produce clearer, more specific, and human-aligned search queries.

\subsection{Reference Discovery}
For each sentence, the system retrieves five supporting references and evaluates performance. Baselines include: (1) Google Scholar search using the raw sentence as the input query, and (2) ChatGPT generation with internet access, prompted to provide five real and relevant references matching the sentence and the surrounding manuscript context. Performance is measured by three metrics:

\begin{itemize}
  \item \textbf{Validity}: percentage of retrieved references that are real and accessible from public academic repositories.
  \item \textbf{Precision}: percentage of retrieved references that are semantically relevant and supportive.
  \item \textbf{Usability}: percentage of cases for which the system returns at least one directly usable reference.
\end{itemize}

\noindent The precision and usability metrics are evaluated through human expert inspection and LLM-as-a-judge using GPT-5.
    
Table~\ref{tab:evaluation} summarizes the performance of three methods on the validation set, showing 100\% valid references with high precision and usability in the proposed system.
An interesting observation emerges from the precision metric: although the LLM-native approach (i.e., ChatGPT) often produces invalid or fabricated references, its precision among valid ones is remarkably high due to the model's strong semantic understanding, enabling it to generate citations closely aligned with the given context and outperform traditional retrieval methods that rely on keyword matching. This explains the generation of non-existent references as a phenomenon to maintain tight semantic alignment with the query.
We further observe a consistent trend where GPT‑5 ratings exceed those of human evaluators, indicating a misalignment in evaluation standards. The discrepancy highlights the need for caution when employing fully autonomous LLM agents in research, as they may compromise scientific rigor, and emphasizes the importance of the LLM‑integrated design implemented in this demonstration system.

\section{Conclusion}

The CiteLLM system represents a significant step toward the trustworthy, context-aware, and privacy-preserving integration of LLMs in academic writing. By automating reference discovery through context-aware query construction, validity verification, and a novel interaction paradigm, researchers can efficiently identify, assess, and cite relevant works without exposing sensitive manuscripts to external servers or fragmented third-party platforms. The proposed approach is hallucination-free, as the LLM assists in retrieval without generating references, aligning with ethical standards for academic integrity. Ultimately, CiteLLM demonstrates how responsible AI design can enhance academic workflows, enabling researchers to engage with scientific knowledge more efficiently while maintaining full control over their intellectual contributions. Future work should focus on developing more user-friendly LLM utilities and optimizing their integration to reduce latency and costs, enabling seamless and impactful AI automation in academic workflows.

\begin{acks}
The work was partially supported by the PolyU Start-up Fund (P0059983), the NSFC/RGC Joint Research Scheme (N\_PolyU5179/25), the National Natural Science Foundation of China (62532002), and the Research Grants Council (Hong Kong) (PolyU25600624).
\end{acks}

\bibliographystyle{ACM-Reference-Format}
\bibliography{references}

\end{document}